\documentclass[10pt,twocolumn,letterpaper]{article}

\usepackage{cvpr}
\usepackage{times}
\usepackage{epsfig}
\usepackage{graphicx}
\usepackage{amsmath}
\usepackage{amssymb}

\usepackage{multirow}
\usepackage{diagbox}

\usepackage[pagebackref=true,breaklinks=true,letterpaper=true,colorlinks,bookmarks=false]{hyperref}

\cvprfinalcopy 

\def\cvprPaperID{0010} 

\ifcvprfinal\fi

\newcommand\blfootnote[1]{%
  \begingroup
  \renewcommand\thefootnote{}\footnote{#1}%
  \addtocounter{footnote}{-1}%
  \endgroup
}
\begin{document}

\title{SUSiNet: See, Understand and Summarize it}

\author{Petros Koutras \quad and \quad Petros Maragos \\
School of E.C.E., National Technical University of Athens, Greece\\
{\tt\small \{pkoutras, maragos\}@cs.ntua.gr}
}


\maketitle

\begin{abstract}
In this work we propose a multi-task spatio-temporal network, called SUSiNet, that can jointly tackle the spatio-temporal problems of saliency estimation, action recognition and video summarization. Our approach employs a single network that is jointly end-to-end trained for all tasks with multiple and diverse datasets related to the exploring tasks. The proposed network uses a unified architecture that includes global and task specific layer and produces multiple output types, i.e., saliency maps or classification labels, by employing the same video input. Moreover, one additional contribution is that the proposed network can be deeply supervised through an attention module that is related to human attention as it is expressed by eye-tracking data. From the extensive evaluation, on seven different datasets, we have observed that the multi-task network performs as well as the state-of-the-art single-task methods (or in some cases better), while it requires less computational budget than having one independent network per each task.
\end{abstract}

\vspace{-0.30cm}
\section{Introduction}

 \begin{figure}[t]
\begin{center}
\includegraphics[width=\linewidth]{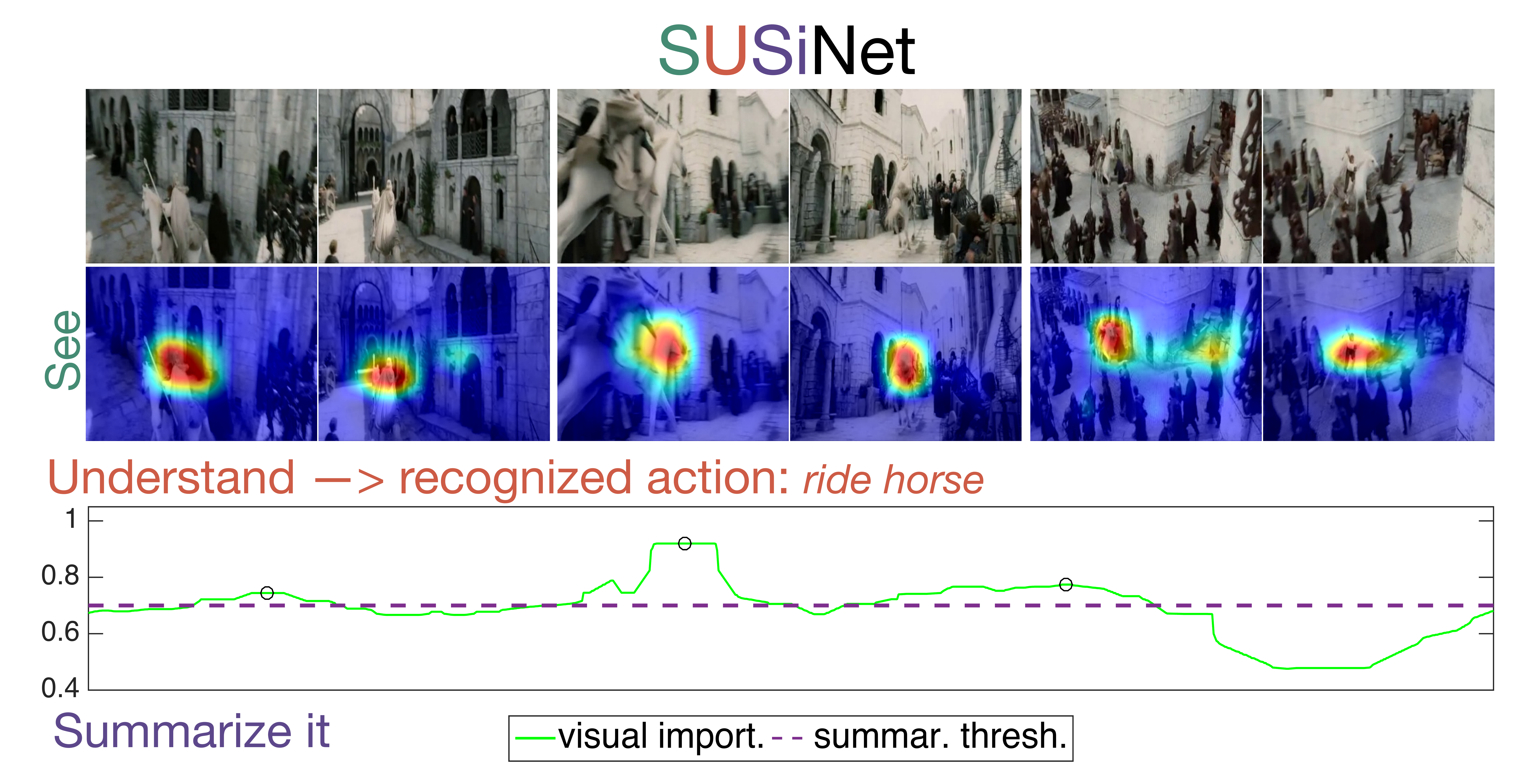}
\end{center}
\vspace{-0.3cm}
   \caption{Example of the multi-task spatio-temporal SUSiNet that can jointly perform \emph{saliency estimation}, \emph{action recognition} and \emph{video summarization} in a similar manner as a human who views the video. The second row depicts the saliency maps produced by our net, which highlight the most important part of the video. In parallel, our network recognizes the visual action ``ride horse'' and computes importance scores (green line) for each frame that indicate which video segments will be included in the summary.}
\label{fig:front_fig}
\vspace{-0.4cm}
\end{figure}
{\blfootnote{{\scriptsize This research has been co-financed by the European Regional Development Fund of the European Union and Greek national funds through the Operational Program Competitiveness, Entrepreneurship and Innovation, under the call Research - Create - Innovate (project code: T1EDK-01248, i-Walk)}} } 
During the last decade, the extensive usage of Convolutional  Neural Networks (CNNs) has boosted the performance throughout  the  majority  of  spatial tasks in computer vision, such as object detection or semantic segmentation \cite{SiZi15,he2016deep,maskRCNN}. Nowadays, automatic video understanding becomes one of the most essential and demanding challenges and research directions due to the increased amount of video data (i.e. YouTube videos, movies, documentaries, home videos). The new problems that have arisen in this field, such as activity recognition, video saliency, scene analysis or video summarization, require the integration and modelling of the temporal evolution instead of applying image based algorithms to each frame independently, such as in video segmentation or pose estimation. In addition, as large-scale video datasets have appeared, an increased performance regarding the video related tasks, i.e., action recognition, can also be achieved \cite{carreira2017quo, hara2018can}. 

However, till recently, the majority of this great progress in computer vision has been achieved by facing each task independently, without trying to jointly solve multiple tasks or investigating relationships between them. Recent works towards this direction have proceeded by training multi-task networks \cite{kokkinos2017ubernet} or finding the structure among visual tasks and apply transfer learning \cite{zamir2018taskonomy}. These approaches open the road for investigating simpler and faster ways to solve multiple task simultaneously, rather than maximize the task-specific performance, and thus providing useful computer vision tools to other scientific communities, i.e., robotics or cognitive sciences.    

In this work we propose the \textbf{S}ee, \textbf{U}nderstand and \textbf{S}ummarize \textbf{i}t Network (SUSiNet) a multi-task spatio-temporal network that can jointly tackle the problems of saliency estimation, visual concept understanding and video summarization. Some examples of visual concepts that often appear in videos are human faces, actions, scenes or objects \cite{Bouritsas2018Multimodal,ray2018scenes}. Let's consider the example that is depicted in Fig.~\ref{fig:front_fig}. During watching a video, first we ``see" the visual information and we focus on the most salient spatio-temporal regions, a process which is related with the visual saliency estimation task. Afterwards, we try to ``understand" what we have seen by recognizing the underlying visual concept, which in this work is approached through the action recognition task. Finally, if someone ask us to ``summarize it" we will think about keeping in the summary the most important parts of the video, which in many cases may include the recognized concept.  

\begin{figure*}[t]
\begin{center}
\includegraphics[width=\textwidth]{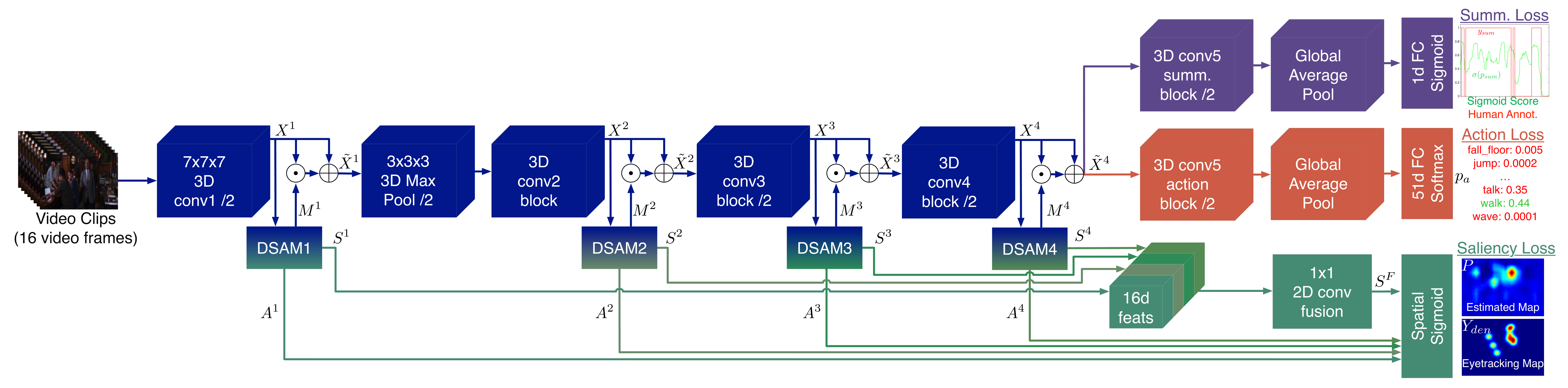}
\end{center}
   \caption{SUSiNet architecture: the multi-task spatio-temporal network is based on the ResNet architecture and has three different branches associated with the different spatio-temporal tasks.}
\label{fig:susi}
\end{figure*}

Our approach employs a single network that is jointly end-to-end trained for all spatio-temporal tasks and incorporates a unified architecture that includes both global and task specific layers. The proposed mutli-task network produces multiple output types, i.e., saliency maps or classification labels, by employing the same video input in the form of 3D video clips. To the best of our knowledge, our approach is the first that deals with these multiple spatio-temporal problems using one single network trained in an end-to-end manner. In \cite{bazzani2016recurrent} the authors have weighted the video clips by the predicted saliency map estimated by their proposed spatio-temporal attention model, in order to compute better feature representations and perform action classification using SVMs classifiers. The work of \cite{ray2018scenes} concatenates deep features, learned independently for the different tasks (scenes, object actions), in order to investigate possible relationships between them. However, none of this approaches employ an end-to-end trained multi-task network for jointly solving multiple spatio-temporal problems simultaneously. Furthermore, one additional contribution is that our new network can be deeply supervised through a proposed attention module that is related to human attention as it is expressed by eye-tracking data. 

For the end-to-end training of the SUSiNet we wish to have available a large scale video database that will contain all the required annotations (i.e., eyetracking data, action labels, human created summaries). 
However, this is not a realistic scenario since the creation of a such dataset requires a lot of human effort for the continuous labelling of a significant amount of videos with all these annotations. So, one important aspect of our proposed network is its ability to be trained with many but diverse datasets, which have been developed for each task independently, following ideas from spatial multi-task networks \cite{kokkinos2017ubernet}. 

We have extensively evaluated SUSiNet performance for all the three tasks on seven different datasets. As we will see the multi-task network performs very close to other state-of-the-art single-task methods that are based on more sophisticated architectures or employ heavyweight inputs \cite{carreira2017quo} (two streams networks and long high-resolution clips) and have been fine-tuned for maximizing a task-specific performance. On the other hand, our approach employs a single network for all tasks and requires less computational budget than having one independent network per each task. 

\section{Related Work}

Since the literature regarding the recognition of the three spatio-temporal tasks we are exploring is quite broad, we refer only some of the most recent deep learning based work about each task. For general reviews about more classical approaches you can see \cite{BoIt13,peng2016bag,Eva+13}.

\noindent\textbf{Visual Saliency:} 
The early CNN-based approaches for saliency were based on the adaptation of pretrained CNN models for visual recognition tasks \cite{Kummerer2014b,vig2014large}. Later, in \cite{Pan_2016_CVPR} both shallow and deep CNN were trained end-to-end for saliency prediction while \cite{huang2015salicon, jetley2016end} trained the networks by optimizing common saliency evaluation metrics. In \cite{Pan_2017_SalGAN} the authors employed end-to-end Generative Adversarial Networks (GAN), while \cite{wang2018deep} has utilized multi-level saliency information from different layer through skip connections. Long Short-term Memory (LSTM) networks have also been used for tracking visual saliency both in static images \cite{cornia2018predicting} and video stimuli \cite{wang2018revisiting}. In order to improve saliency estimation in videos, many approaches employ multi-stream networks, such as RGB/Optical Flow (OF) \cite{bak2017spatio}, RGB/OF/Depth \cite{leifman2017learning}, or multiple subnets such as objectness/motion \cite{jiang2018deepvs} or saliency/gaze \cite{Gorji_2018_CVPR} pathways.

\noindent\textbf{Action Recognition:} The work of \cite{karpathy2014large} explored several approaches for fusing information over temporal dimension, while in \cite{ji2013} 3D spatio-temporal convolutions have been proposed, whose performance can be boosted when trained on large datasets \cite{tran2015learning,varol2018long} or employing ResNet architectures \cite{hara2018can}. Recently, many approaches have tried to separate the spatial and temporal parts of 3D convolutions \cite{sun2015human,Wang_2018_CVPR,tran2018closer}, which has achieved improvements over the conventional 3D networks. The work of \cite{simonyan_two-stream_2014} introduced the two-stream networks that employ RGB and optical flow inputs for modelling the appearance and motion information. This framework has become the basis for many other methods \cite{Wang_TDD_2015,feichtenhofer2016,feichtenhofer_resconvnets_2016,wang2016temporal,chen_ssn_2017,feichtenhofer2017spatiotemporal}. In \cite{carreira2017quo} the two-stream networks were combined with 3D convolutions, forming the I3D model that holds the state-of-the-art performance on the most action datasets.    

\noindent\textbf{Video Summarization:} Many of the recent methods approach video summarization as an optimization problem \cite{GygliCVPR15,song2015tvsum,zhang2016summary,elhamifar2017online,plummer2017enhancing}, or they employ recurrent neural networks \cite{zhang2016video, zhao2018hsa}, i.e., LSTMs, in order to model long-time dependencies inside the video which are very important in summary creation. In \cite{mahasseni2017unsupervised} the authors proposed a summarization framework based on GANs, where generator consists of an LSTM-based autoencoder and the discriminator is another LSTM. The work of \cite{zhang2018retrospective} combines sequential models for the summary creation with a retrospective encoder which maps the summaries to an abstract semantic space. Following a different framework, the authors of \cite{rochan2018video}, inspired by the progress in semantic segmentation, proposed fully convolutional sequence networks as an alternative approach to recurrent networks.

\section{Multi-task Spatio-Temporal Network}

The proposed spatio-temporal network deals with three different tasks simultaneously and produces different types of outputs by employing the same video input in the form of small video clips. For the saliency estimation, where we face a spatial estimation problem, the network output consists of a saliency map, while for the action classification task we have the classical softmax scores. For the video summarization we need to estimate video's segments importance scores, which indicate whether a segment will be included in the summary, as the sigmoid scores of a binary classification problem.     

\subsection{Global Architecture with Deep Supervision}
The whole architecture of our multi-task network, which is shown in Fig.~\ref{fig:susi}, is based on the general ResNet architecture \cite{he2016deep} and specifically the 3D extension proposed in \cite{hara2018can} for the problem of action classification. The global pathway of the network with parameters $\mathbf{W}_{GL}$ (dark blue), which is shared among all the tasks, includes the first four convolutional blocks $\mathrm{conv1, conv2, conv3, conv4}$ from the employed ResNet version that provides outputs $X^m, m=1,\ldots,4$ in different spatial and temporal scales. In order to enhance the most salient regions of these feature representations, we apply an attention mechanism by taking the element-wise product between each channel of the feature map $X^m$ and the attention map $M^m$:     
\begin{equation}
\tilde{X}_m = (1+M^m)\odot X_m, \quad m=1,\ldots,4.
\end{equation}
The attention map is obtained by our proposed Deeply Supervised Attention Module (DSAM) based on the idea of deep supervision that has been used in edge detection \cite{xie2015holistically}, object segmentation \cite{Cae+17} and static saliency \cite{wang2018deep}. In contrary to these previous works the proposed module is used for both enhancing the feature representations of the global network as well as providing the multi-level saliency maps for the task of spatio-temporal saliency. Thus, the DSAM parameters $\mathbf{W}_{AM}^m$ are trained by both the main-path of the network, which is shared among all the tasks, and the eye-tracking data that are used for the task of saliency estimation through the skip connections of the Fig.~\ref{fig:susi}. In this way, we enrich our network with an attention module that is related to human attention as it is expressed by eye-tracking data.

\begin{figure}[t]
\begin{center}
\includegraphics[width=0.45\textwidth]{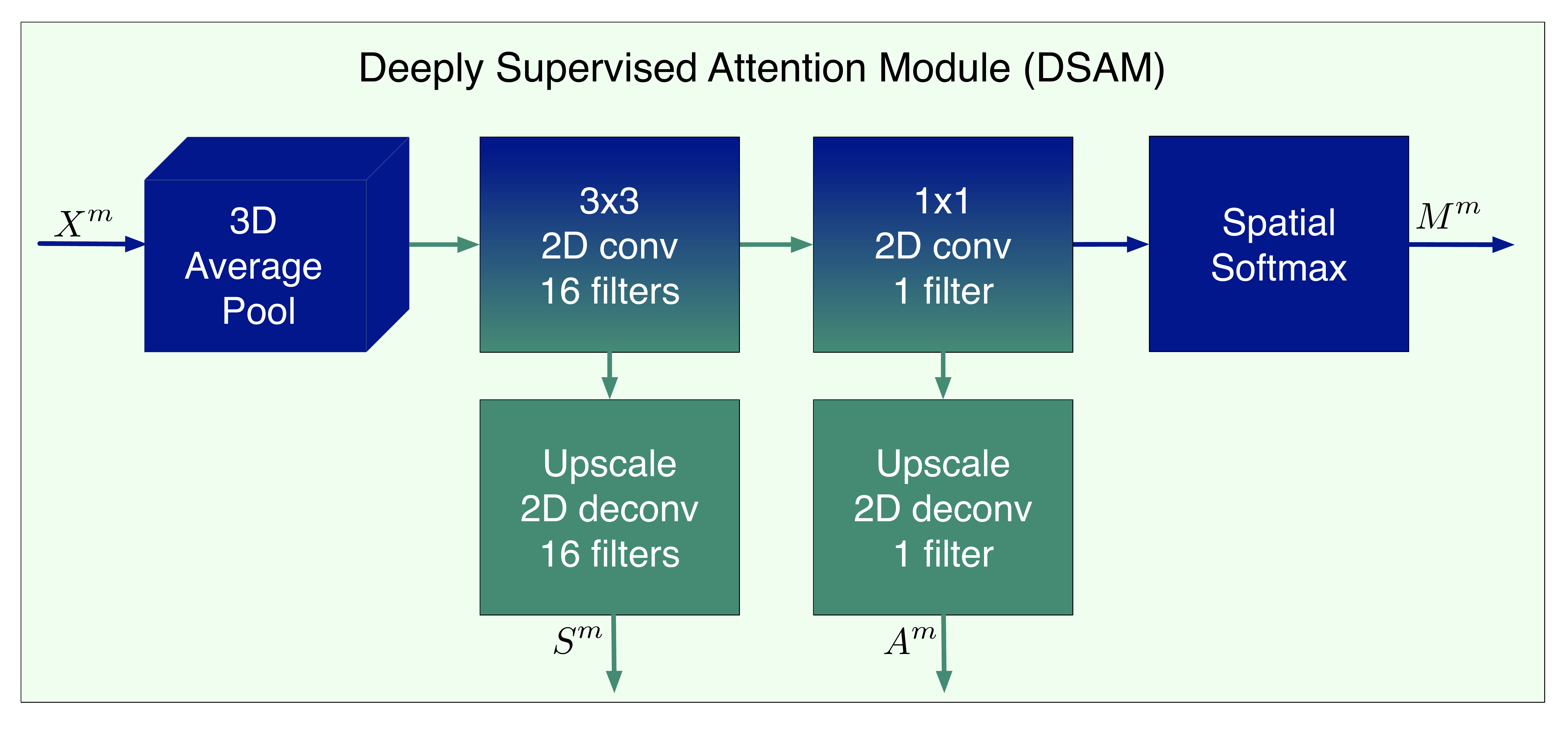}
\end{center}
   \caption{Deeply Supervised Attention Module (DSAM) enhances the global network's representations and provides the multi-level saliency maps for the task of spatio-temporal saliency.}
\label{fig:dsam}
\end{figure}

Figure~\ref{fig:dsam} shows the architecture of the attention module applied at level $m$. It includes an averaging pooling in the temporal dimension followed by two spatial convolution layers that provide the saliency features $S^m$ and the activation map $A^m$. Both of these representations are up-sampled (using the appropriate deconvolution layers) to the initial image dimensions and used for the deep supervision of the module as well as for the multi-level saliency estimation. The attention map $M^m(x,y)$ is given through a spatial softmax operation applied at the activation map $A^m(x,y)$:
\begin{equation}
M^m(x,y) = \frac{\exp(A^m(x,y))}{\sum_x \sum_y \exp(A^m(x,y))}.
\end{equation}

\subsection{Task-specific Sub-Networks}

\subsubsection{Visual Saliency Module}
Since in the saliency estimation we face a dense prediction problem, we need to employ a fully convolution sub-network with parameters $\mathbf{W}_{sal}$ (see green parts of Fig.~\ref{fig:susi}) that takes advantage from the concatenated multi-level saliency features $S^m$ of the DSAM components and  produces the final fused saliency map $S^F$ which corresponds to the given video clip. For the training of the network parameters $\mathbf{W'}_{sal}=[\mathbf{W}_{sal}, \mathbf{W}_{AM}^{1:4}, \mathbf{W}_{GL}]$, which are associated with visual saliency, the deep attention supervision of the whole multi-task network and the global branch we construct a loss that compares the saliency map $S^F$ and the activations $A^m$ with the ground truth maps $Y_{sal}$ obtained by the eye-tracking data:
\begin{equation}
\begin{split}
\mathcal{L}_{sal}(\mathbf{W'}_{sal}) = \mathcal{D}(\mathbf{W'}_{sal} | \sigma(S^F),Y_{sal}) + \\ 
\sum_{m=1}^4 \mathcal{D}(\mathbf{W}_{AM}^m | \sigma(A^m),Y_{sal}), \label{sal_loss_gen}
\end{split}
\end{equation}
where $\sigma(\cdot)$ denotes the sigmoid non-linearity and $\mathcal{D}(\cdot)$ is a loss function between the estimated and the ground truth 2D maps. In the saliency evaluation several different metrics are employed in order to compare the predicted saliency map $P \in [0,1]^{N_X \times N_Y}$ with the eyetracking data \cite{bylinskii2016different}. As ground truth maps we are using either the map of fixation locations $Y_{fix} \in \{0,1\}^{N_X \times N_Y}$ on the image plane of size $N_X \times N_Y$ or the dense saliency map $Y_{den}\in[0,1]^{N_X \times N_Y}$, which arises by convolving the binary fixation map with a gaussian kernel. Thus, as $\mathcal{D}(\cdot)$ we employ three loss functions associated with the different aspects of saliency evaluation. The first is the cross-entropy loss between the predicted map $P$ and the thresholded dense map $\tilde{Y}_{den}$:
\begin{equation}
\begin{split}
\mathcal{D}_{CE}(\mathbf{W}|P,\tilde{Y}_{den})= - \sum_{x,y} \tilde{Y}_{den}(x,y) \odot \log(P(x,y;\mathbf{W})) \\
+ (1-\tilde{Y}_{den}(x,y)) \odot (1-\log(P(x,y;\mathbf{W}))).
\end{split}
\end{equation}
In order to handle the strong imbalance between the salient and non-salient pixels we take a variant of the above loss, which has been effectively used in other imbalanced tasks as boundrary detection \cite{xie2015holistically,kokkinos2015pushing,maninis2016convolutional}: 
\begin{equation}
\begin{split}
\tilde{\mathcal{D}}_{CE}(\mathbf{W}|P,\tilde{Y}_{den})= - \beta \cdot \sum_{x,y \in \mathcal{Y}_+} \log(P(x,y;\mathbf{W})) \\
- (1-\beta) \cdot \sum_{x,y \in \mathcal{Y}_-} (1-\log(P(x,y;\mathbf{W}))),  
\end{split}
\end{equation}
where $\mathcal{Y}_+$, $\mathcal{Y}_-$ are the set of salient and non-salient pixels respectively and $\beta=|\mathcal{Y}_-|/(|\mathcal{Y}_+| + |\mathcal{Y}_-|)$.
The second employed loss function is based on the linear Correlation Coefficient (CC) that is widely used in saliency evaluation and measures the linear relationship between the predicted saliency $P$ and the dense ground truth map $Y_{den}$:
\begin{equation}
\mathcal{D}_{CC}(\mathbf{W}|P,Y_{den}) = -\frac{\mathrm{cov}(P(x,y;\mathbf{W}),Y_{den}(x,y))}{\rho(P(x,y;\mathbf{W}))\cdot\rho(Y_{den}(x,y))}, 
\end{equation} 
where $\mathrm{cov},\rho$ denote the covariance and the standard deviation respectively. 

The last loss is derived from the Normalized Scanpath Saliency (NSS) metric, which is computed
as the estimated map values $\tilde{P}(x,y;\mathbf{W})=\frac{P(x,y;\mathbf{W})-\mu(P(x,y;\mathbf{W}))}{\rho(P(x,y;\mathbf{W}))}$, after zero mean normalization and unit standardization, at human fixation locations ($Y_{fix}(x,y)=1$):
\begin{equation}
\mathcal{D}_{NSS}(\mathbf{W}|\tilde{P},Y_{fix}) = - \frac{1}{N_{f}} \sum_{x,y} \tilde{P}(x,y;\mathbf{W})\odot Y_{fix}(x,y),
\end{equation}
where $N_{f}=\sum_{x,y}Y_{fix}(x,y)$ denotes the total number of fixation points. 

The final loss of the $j$-th input sample for the task of visual saliency estimation is given by a weight combination of the losses $\mathcal{L}^j_{CE}, \mathcal{L}^j_{CC}, \mathcal{L}^j_{NSS}$, which are given by (\ref{sal_loss_gen}) using the corresponding loss functions $\mathcal{D}^j_{CE}, \mathcal{D}^j_{CC}, \mathcal{D}^j_{NSS}$:
\begin{equation}
\mathcal{L}_{sal}^j(\mathbf{W'}_{sal}) = w_{1}\mathcal{L}^j_{CE} + w_{2}\mathcal{L}^j_{CC} + w_{3}\mathcal{L}_{NSS}^j,
\end{equation}  
where $w_1,w_2,w_3$ are the weights of each loss type.

\subsubsection{Action Recognition Module}

For the action recognition problem, which constitutes a classical multi-class problem, we build the task specific layers (with parameters $\mathbf{W}_{act}$) after the output $\tilde{X}_4$ of the global branch. As we can see from Fig.~\ref{fig:susi} (orange blocks), we have a 3D convolutional block, which has identical structure as the $conv5$ block of the employed ResNet architecture, a global average pooling across the temporal dimension and a $C_a$-dimension fully connected layer, where $C_a$ is the number of classes. For the training of action-related parameters $\mathbf{W'}_{act}=[\mathbf{W}_{act}, \mathbf{W}_{AM}^{1:4}, \mathbf{W}_{GL}]$ we employ the standard multi-class cross-entropy for the softmax activations $p_a(c;\mathbf{W}'_{act}),\:c=1,\ldots, C_a$ of the final layer:
\begin{equation}
\mathcal{L}^j_{act}(\mathbf{W}'_{act})=-\log p_a^j(c_j;\mathbf{W}'_{act}),
\end{equation}
where $p_a^j, c_j$ denote the activation and the ground truth class of the $j$-th input sample respectively.

\subsubsection{Summarization Module}

Regarding the summarization task we employ a sub-network with parameters $\mathbf{W}_{sum}$ (Fig.~\ref{fig:susi} - purple blocks) that has a similar structure with the one we used for action recognition, with the difference that the last fully connected layer has only one dimension since we have a binary classification problem (important vs. non-important video segments). The importance score of each video clip is given by the sigmoid activation of the final full-connected layer $\sigma(p_{sum}) \in [0,1]$, while for the training of the all task-related parameters $\mathbf{W'}_{sum}=[\mathbf{W}_{sum}, \mathbf{W}_{AM}^{1:4}, \mathbf{W}_{GL}]$ we employ the binary cross-entropy (BCE) loss. Since in most annotated databases only a small portion of the whole video is annotated as important and selected for final summary, the ground truth data are heavily biased. Thus we use a weighted variant of the BCE based on the ratio $\gamma=\frac{|\mathcal{S}_-|}{|\mathcal{S}_+|}$ between the number of negative and positive samples in the whole training dataset:
\begin{equation}
\begin{split}
\mathcal{L}^j_{sum}(\mathbf{W}'_{sum}) = 
- \gamma \cdot y_{sum}^j \cdot \log(\sigma(p_{sum}^j(\mathbf{W}'_{sum})) \\
- (1-y_{sum}^j) \cdot (1-\log(\sigma(p_{sum}^j(\mathbf{W}'_{sum}))),
\end{split}
\end{equation}
where $y_{sum}^j \in [0,1]$ denotes the ground truth annotation regarding the importance of the $j$-th video clip.   

\subsection{Multi-task Training}

For the end-to-end training of the whole multi-task spatio-temporal network ($\mathbf{W}_{all}=[\mathbf{W}_{GL},\mathbf{W}_{AM}^{1:4},\mathbf{W}_{sal},\mathbf{W}_{act},\mathbf{W}_{sum}]$) we can simply minimize the sum of the above task-specific losses over all the samples of the batch $\mathcal{B}$:
\begin{eqnarray}
\begin{split}
&\mathcal{L}(\mathbf{W}_{all}) = \alpha_{sal} \sum_{j \in \mathcal{B}} \mathcal{L}_{sal}^j(\mathbf{W}'_{sal}) \\
&+ \alpha_{act} \sum_{j \in \mathcal{B}} \mathcal{L}_{act}^j(\mathbf{W}'_{act}) 
+ \alpha_{sum} \sum_{j \in \mathcal{B}} \mathcal{L}_{sum}^j(\mathbf{W}'_{sum}),
\end{split}
\end{eqnarray}
where $\alpha_{sal},\alpha_{act},\alpha_{sum}$ are weights that control the contribution of each task. This approach, which has been followed in many static multi-task networks \cite{dai2016instance,eigen2015predicting,gkioxari2015contextual}, assumes that each sample of the batch has annotations for all tasks. However, as \cite{kokkinos2017ubernet} has mentioned, this is not a realistic scenario, especially for our tasks where none of the annotations can be derived  from the other tasks' annotations, as in object detection and semantic segmentation. Thus, we use the Asynchronous Stochastic Gradient Descent (SGD) algorithm, which has been proposed in \cite{kokkinos2017ubernet}, that allows us to have different effective batchsizes $\mathcal{B}_{sal}, \mathcal{B}_{act}, \mathcal{B}_{sum}$ and update the parameters of the task-specific layers once we have seen enough samples. The updates for the shared parameters $W_{GL}'=[W_{GL} \mathbf{W}_{AM}^{1:4}]$  of the multi-task network are based on the sum of the gradients from all losses:
\begin{flalign}
&\mathbf{d}\mathbf{W}_{GL}' = \sum_{j \in \mathcal{B}}  \alpha_{sal} \nabla_{\mathbf{W}_{GL}'}\mathcal{L}_{sal}^j(\mathbf{W}'_{sal})  \\
&+ \alpha_{act} \nabla_{\mathbf{W}_{GL}'}\mathcal{L}_{act}^j(\mathbf{W}'_{act})  
+ \alpha_{sum}  \nabla_{\mathbf{W}_{GL}'}\mathcal{L}_{sum}^j(\mathbf{W}'_{sum}), \nonumber
\end{flalign} 
where $\mathcal{B}=\mathcal{B}_{sal} \cup \mathcal{B}_{act} \cup \mathcal{B}_{sum}$ is the total minibatch that contains all the training samples. The updates for the task-specific parameters depend only on the gradient of each different loss:
\begin{equation}
\begin{split}
\mathbf{d}\mathbf{W}_{sal} &= \sum_{j \in \mathcal{B}_{sal}}  \alpha_{sal} \nabla_{\mathbf{W}_{sal}}\mathcal{L}_{sal}^j(\mathbf{W}'_{sal}) \\
\mathbf{d}\mathbf{W}_{act} &= \sum_{j \in \mathcal{B}_{act}}  \alpha_{act} \nabla_{\mathbf{W}_{act}}\mathcal{L}_{act}^j(\mathbf{W}'_{act}) \\
\mathbf{d}\mathbf{W}_{sum} &= \sum_{j \in \mathcal{B}_{sum}}  \alpha_{sum} \nabla_{\mathbf{W}_{sum}}\mathcal{L}_{sum}^j(\mathbf{W}'_{sum})
\end{split}
\end{equation}

 \begin{table*}[th!]
 \centering{
 \resizebox{0.98\textwidth}{!}{
 \begin{tabular}{ l | c c c c || c c c c || c c c c}
 \hline
 \multirow{2}{*}{\backslashbox{ \kern-0.5em Method \kern-0.5em}{\kern-1.9em Dataset \kern-0.5em}} &  \multicolumn{4}{c||}{DIEM}   &  \multicolumn{4}{c||}{DFK1K} & \multicolumn{4}{c}{ETMD} \\ \cline{2-13} 
 & CC $\uparrow$ & NSS $\uparrow$ & AUC-J $\uparrow$ & sAUC $\uparrow$ & CC $\uparrow$ & NSS $\uparrow$ & AUC-J $\uparrow$ & sAUC $\uparrow$ & CC $\uparrow$ & NSS $\uparrow$ & AUC-J $\uparrow$ & sAUC $\uparrow$ \\ \hline \hline
\textbf{SUSiNet (1-task)} [ST] & \textbf{0.6138} & \textbf{2.4267} & \textbf{0.8736} & \textbf{0.6747} & \textbf{0.4676} & \textbf{2.5908} & \textbf{0.8843} & \textbf{0.6991} & \textbf{0.5523} & \textbf{2.8365} & \textbf{0.9173} & \textbf{0.7312} \\ 
\textbf{SUSiNet (multi)} [ST] & \textbf{0.5614} & \textbf{2.1398} & \textbf{0.8810} & \textbf{0.6736} & \textbf{0.4116} & \textbf{2.2092} & \textbf{0.8910} & \textbf{0.6980} & \textbf{0.4780} & \textbf{2.3642} & \textbf{0.9162} & \textbf{0.7272} \\ \hline \hline
Deep-Net \cite{Pan_2016_CVPR} [S] & 0.4305 & 1.6238 & 0.8401 & 0.6262 & 0.2969 & 1.5804 & 0.8421 & 0.6432 & 0.3438 & 1.6523 & 0.8712 & 0.6588 \\
DVA \cite{wang2018deep} [S] & 0.5179 & 2.1607 & 0.8599 & 0.6400 & 0.3593 & 2.0644 & 0.8609 & 0.6572 & 0.4228 & 2.2507 & 0.8848 & 0.6843 \\
SAM \cite{cornia2018predicting} [S] & 0.5352 & 2.2482 & 0.8651 & 0.6429 & 0.3684 & 2.1180 & 0.8680 & 0.6562 & 0.4345 & 2.3155 & 0.8890 & 0.6875 \\ 
ACLNet \cite{wang2018revisiting} [ST] & 0.5626 & 2.2168 & 0.8717 & 0.6228 & 0.4167 & 2.2962 & 0.8883 & 0.6523 & 0.4508 & 2.2058 & 0.9073 & 0.6482 \\
DeepVS \cite{jiang2018deepvs} [ST] & 0.4885 & 2.0352 & 0.8448 & 0.6248 & 0.3500 & 1.9680 & 0.8561 & 0.6405 & 0.4316 & 2.3030 & 0.8955 & 0.6672 \\ \hline 
\end{tabular}
}}
\vspace{0.2cm}
\caption{Evaluation results for the visual saliency estimation task. In most cases, the proposed multi-task SUSiNet outperforms the existing state-of-the-art methods for video saliency over all three different datasets according the four evaluation metrics. [ST] stands for spatio-temporal models while [S] denotes a spatial only model that is applied to each frame independently.}
\label{table:sal_eval}
\end{table*}

\subsection{Implementation}

Our implementation and experimentation with the proposed multi-task network uses as backbone the 3D ResNet-50 architecture \cite{hara2018can} that has showed competitive performance against other deeper architectures for the task of action recognition in terms of performance and computational budget. As starting point we used the weights from the pretrained model in the Kinetics 400 database. 

\noindent \textbf{Training:} For the training we used the asynchronous version of stochastic gradient descent with momentum 0.9 while we also assign a weight decay of 1e-5 for regularization. We have also employed effective batchsizes of 128 samples for all tasks while the learning rate has started from 0.01 and divided by 10 when the loss saturated. 
The weights $w_1, w_2, w_3$ for the saliency loss are selected 0.1, 2, 1 after experimentation, while the ratio $\gamma$ in the summarization loss was set to 3.06 based on the statistics of the employed training datasets. The weights $\alpha_{sal},\alpha_{act},\alpha_{sum}$, which control the importance of each task, have been experimentally tuned to 0.1, 1, 1 (based on the losses' ranges) in order to avoid the overfitting of the network to one task.

\noindent \textbf{Data Augmentation:} The input samples in the network consist of 16-frames RGB video clips spatially resized at $112 \times 112$ pixels. We have also applied data augmentation for random generation of training samples. For the action recognition task we randomly sampled a 16-frame clip from each training video and afterwrds we followed the procedure of random multi-scale spatial cropping and flipping, which is described in \cite{wang2016temporal}. For the summarization task we divided the initial long-duration videos into 90-frames non-overlapping segments and generated the 16-frames clip following the same procedure as in action recognition task. Regarding the human annotations we took its average inside the created clips, that gave training samples with slightly different annotation scores and helped us to avoid the network's overfitting.    For data augmentation in the saliency estimation task, we followed a similar approach as in summarization task without the random cropping step. We applied the same spatial transformations to the 16 frames of the video clip and the eye-tracking based saliency maps of the median frame, which has been considered as the the ground truth map of the whole clip.

\noindent \textbf{Testing:} During the testing phase, for the action recognition task we extracted the network predictions using a 16-frames non-overlapping sliding window where each clip is spatially cropped at the center position with scale 1. Then, we computed the final action label for each video by simply averaging the clips' predictions, while for the summarization task we took frame-wise importance scores by repeating the values of the 16-frame clips' scores. Finally, for saliency estimation we obtained an estimated saliency map per frame using a 16-frame sliding window with step 1 without any spatial cropping.     

\section{Multiple Tasks Evaluation}

\subsection{Datasets}

For the training and the evaluation of the proposed multi-task network we wish to have a large scale video database that will contain eyetracking annotation, labelling of the performed actions as well as continuous human annotation of the frames importance or equivalently human created summaries. However, this is not a realistic scenario since many datasets have been developed for each task but none of them contains all of the three required types of annotation. Very recently, \cite{ray2018scenes} proposed a multi-task and multi-label video dataset aiming to the recognition of different visual concepts (scenes, objects, actions) which are different from our investigated tasks. Note that since our multi-task network is modular, it could be extended to recognize and understand more visual concepts such as objects or scenes. 

The most relevant dataset to our tasks is the \mbox{COGNIMUSE} database \cite{ZKE+_review, cogn_data}, which constitutes a video database annotated with ground-truth annotations for frame-wise sensory and semantic importance as well as  audio and visual events. It is a generic database that has been used for video summarization \cite{CVSP_ICIP2015}, as well as audio-visual concept recognition \cite{Bouritsas2018Multimodal}. The creators of the database have also developed the Eye-Tracking Movie Database (ETMD) \cite{koutras2015perceptually}, which contains eyetracking annotations for a subset of the \mbox{COGNIMUSE} videos. For our experiments we have used the 30-minutes excerpts from the seven movies \footnote{``A Beautiful Mind'' (BMI), ``Gladiator'' (GLA), ``Chicago'' (CHI), ``Finding Nemo'' (FNE), ``Lord of the Rings - the Return of the King'' (LOR), ``Crash'' (CRA), ``The Departed'' (DEP)} as well as the full movie ``Gone With the Wind'' (GWW) that they have at least two of the three annotation types (see Table~\ref{table:cogn_eval}). For the training we followed an 8-fold (leaving one movie out) cross-validation approach.

One important aspect of the proposed multi-task network is its ability to be trained with diverse datasets. So, we employ five more state-of-the-art datasets, containing annotations only for a specific task in order to increase the training set as well as compare our results with other state-of-the-art methods. Specifically, for the visual saliency estimation we employ the DIEM dataset \cite{mital2011clustering}, which contains eyetracking data for 84 videos with duration between 27-217 sec from 50 observers, and the DFK1K \cite{wang2018revisiting}, with eyetracking data from 17 observers over 1000 videos with duration 17-42 sec. During the experiments, for the DIEM we followed the ``train-test" split of \cite{BSI13}, while for the DFK1K we used the validation set for our testing since the test set is not publicly available. Regarding the action recognition task we employ the HMDB51 dataset \cite{kuehne2011hmdb} that includes 6766 video from 51 human action classes. We decided to use this additional dataset because its classes have also been included in the \mbox{COGNIMUSE} dataset. Finally, for the summarization task we used the SumMe \cite{gygli2014creating} and the TVSum50 \cite{song2015tvsum} datasets that include 25 (1.5 to 6.5 minutes length) and 50 (1 to 5 minutes length) videos respectively, mainly from YouTube resources. For the experiment that involves these datasets we have followed a 5-fold cross validation.   

\subsection{Experimental Results}

For the evaluation of the multi-task network we have constructed two types of experiments. In the first, which we refer as ``SUSiNet (1-task)" we trained our network independently for each task using only the task-related datasets. In the ``SUSiNet (multi)" we trained the multi-task network jointly for all the three task employing all the available datasets. Next, we evaluate our results for each different task and compare them with several methods that have achieved state-of-the-art performance for each task independently.  

\begin{figure*}[t]
\begin{center}
\includegraphics[width=0.9\textwidth]{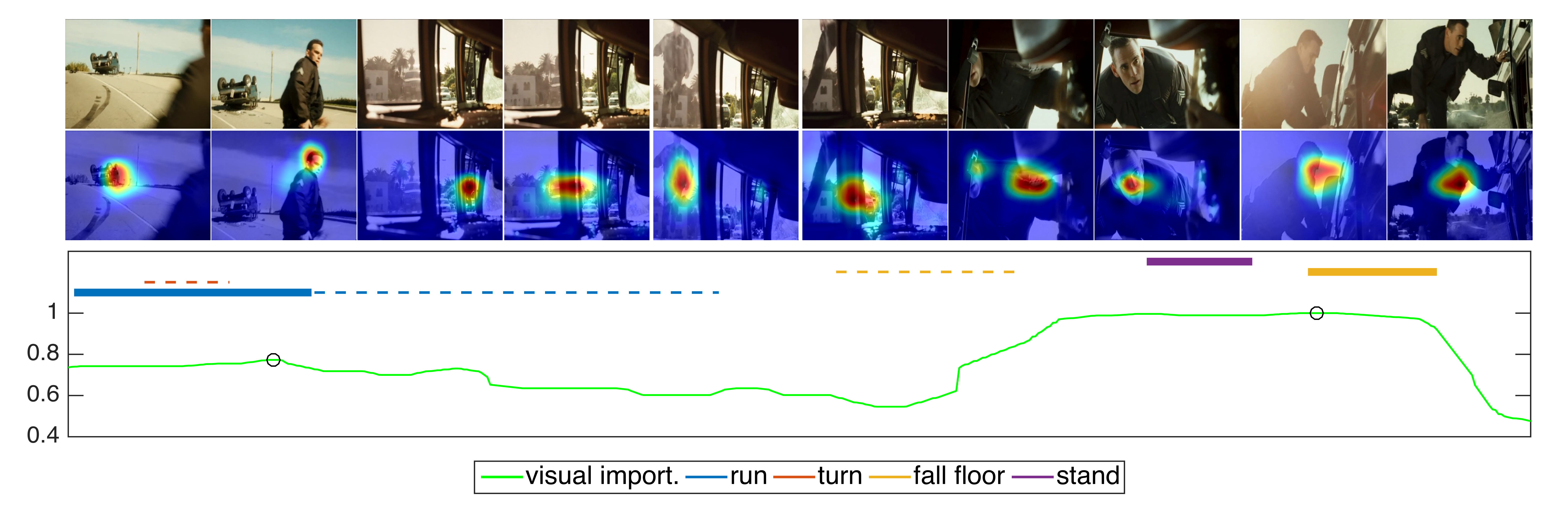}
\end{center}
\vspace{-0.5cm}
   \caption{Example results of the proposed multi-task SUSiNet for the three different tasks: saliency estimation, action recognition, video summarization. The second row consists of the estimated saliency maps, while in the figure we also see the action recognition results (with dotted lines the annotated actions that are not correctly recognized) and the green curve that indicates the frame-wise importance score.}
\label{fig:susi_example}
\end{figure*}

\noindent\textbf{Saliency Estimation Evaluation:} In Table~\ref{table:sal_eval} we present the evaluation results of the proposed SUSiNet on the 3 different datasets and compare its performance against 5 state-of-the-art methods (using their publicly available codes). We employed four widely-used evaluation metrics \cite{bylinskii2016different}: CC, NSS, AUC-Judd (AUC-J) and shuffled AUC (sAUC). In the sAUC we have selected the negative samples from the union of all viewers' fixations across all other frames except the frame for which we compute the AUC. As we see, our method outperforms all the other methods over all the datasets according to all employed metrics. Note there is a small decrease in the CC and NSS scores of the multi-task network compared to the single-task, while according to AUC-based metrics the multi-task network performs in equally or better than the single one. Moreover, the SUSiNet, which is based on 3D spatio-temporal convolutions, achieves to outperform other spatio-temporal methods (ACLNet, DeepVS) that  rely on the LSTM based modelling of the visual saliency. In Figures~\ref{fig:front_fig},~\ref{fig:susi_example} we see examples of our saliency predictions, which in most cases are focused on humans or actions.

\begin{table}[t]
 \resizebox{\linewidth}{!}{
 \begin{tabular}{ l | c | c  }
 \hline
 Method & Pre-train dataset & Aver. Accuracy \\ \hline
  \textbf{SUSiNet (1-task)} & Kinetics & \textbf{60.2} \\
 \textbf{SUSiNet (multi)} & Kinetics & \textbf{62.7} \\ \hline \hline
 C3D \cite{tran2015learning} & Sports-1M & 51.6 \\
 3D ResNet-18 \cite{hara2018can} & Kinetics & 56.4 \\
 3D ResNet-50 \cite{hara2018can} & Kinetics & 61.0 \\
 3D ResNeXt-101 \cite{hara2018can} & Kinetics & 63.8 \\ 
  RGB I3D (64f) \cite{carreira2017quo} & ImageNet \& miniKinetics & 66.4 \\ \hline \hline
 \multicolumn{3}{c}{Other Methods} \\ \hline
 $\mathrm{F_{ST}CN}$ \cite{sun2015human} & ImageNet & 59.1 \\
 ARTNet \cite{Wang_2018_CVPR} & Kinetics & 67.6 \\
 R(2+1)D-RGB \cite{tran2018closer} & Kinetics & 74.5 \\ \hline
 Two-Stream TSN \cite{wang2016temporal} & ImageNet & 68.5 \\
 Two-Stream STM-Nets \cite{feichtenhofer2017spatiotemporal} & ImageNet & 68.9 \\
 Two-Stream I3D (64f)\cite{carreira2017quo} & ImageNet \& Kinetics & 80.7 \\
 Two-Stream R(2+1)D \cite{tran2018closer} & Kinetics & 78.7  
 \\ \hline
\end{tabular}
}
\vspace{0.1cm}
\caption{Evaluation results for the action recognition task over all splits of HMDB51. The proposed multi-task SUSiNet outperforms the single-task as well as several state-of-the-art methods that rely on 3D convolutions. 
}
\label{table:action_eval}
\end{table}

\noindent\textbf{Action Classification Evaluation:} In Table~\ref{table:action_eval} we see the evaluation results of our method on the HMDB51. We can observe that the multi-task network achieves better performance than the single-task. Comparing our method against several other approaches, which are based on 3D CNN networks, we see that our network performs better than the most of them. Note that our network is based on ResNet-50 architecture and uses 16-frames inputs, while 3D ResNeXt or I3D are employing more complex networks or longer clips. For completeness we also report several other methods from literature, which are based on techniques for decoupling the spatial and temporal parts of 3D convolutions or employ two-streams (RGB, Optical Flow) networks, that are not directly compared with our method. However, our proposed network is modular and can be modified and extended to include such techniques but we leave this direction for future work as the scope of this paper is to propose a multi-task spatio-temporal network rather than achieve the best performance for each single task. 

Finally, in Table~\ref{table:cogn_eval} we report our method's results for action recognition in \mbox{COGNIMUSE} database, where we see again that the multi-task network slightly outperforms the single-task. Regarding the lower recognition scores (comparing with HMDB51), we have observed that \mbox{COGNIMUSE} dataset contains many background actions or supplementary actions that may overlap with a main action (i.e., in Fig.~\ref{fig:susi_example} the action ``turn" overlaps with the ``run") and thus it constitutes a very challenging dataset.

\begin{table}[!tb]
 \centering{
 \resizebox{0.84\linewidth}{!}{
 \begin{tabular}{ l | c | c  }
 \hline
 Method & SumMe (F-score) & TVSum50 (F-score) \\ \hline
 \textbf{SUSiNet (1-task)} & \textbf{41.10} & \textbf{59.20} \\
 \textbf{SUSiNet (multi)} & \textbf{40.80} & \textbf{57.00} \\ \hline \hline
 vsLSTM \cite{zhang2016video} & 37.6 [41.6] & 54.2 [57.9]   \\
 HSA-RNN \cite{zhao2018hsa} & 44.1 & 59.8 \\
 SEQ2SEQ \cite{zhang2018retrospective} & 40.8 & 56.3 \\ 
 SUM-FCN \cite{rochan2018video} & 47.5 [51.1]  & 56.8 [59.2]  \\ \hline \hline
 \multicolumn{3}{c}{Other Methods} \\ \hline
 Gygli et al. \cite{GygliCVPR15} & 39.7 & - \\
 dppLSTM \cite{zhang2016video} & 38.6 [42.9]  & 54.7 [59.6]  \\
 SUM-GAN \cite{mahasseni2017unsupervised} & 41.7 [43.6]  & 56.3 [61.2]  \\
 re-SEQ2SEQ \cite{zhang2018retrospective} & [44.9] & [63.9] 
 \\ \hline
\end{tabular}
}}
\vspace{0.2cm}
\caption{Evaluation results for the video summarization task over the SumMe and TVSum50 datasets. 
}
\label{table:sum_eval}
\end{table}

\begin{table}[!tb]
\centering{
 \resizebox{0.95\linewidth}{!}{
 \begin{tabular}{l||c c|c c|c c}
 \hline
 Task & \multicolumn{2}{c|}{Saliency (sAUC)} & \multicolumn{2}{c|}{Action (Acc.)} & \multicolumn{2}{c}{Summar. (AUC)}            
 \\ \hline
 SUSiNet & 1-task & multi & 1-task & multi & 1-task & multi \\ \hline
 BMI & -  	    & -        		   & \textbf{51.54} & 49.88 & 0.7831 & \textbf{0.8023}
 \\ 
 GLA & \textbf{0.6859}  & 0.6727    & \textbf{48.92} & 46.77 & \textbf{0.7863} & 0.7843
 \\ 
 CHI & \textbf{0.7601}  & 0.7565    & 49.41 & \textbf{50.82} & \textbf{0.7901} & 0.7826
 \\ 
 FNE & 0.7224  & \textbf{0.7236}    & -     & -     			&  \textbf{0.5490} & 0.5306 
 \\ 
 LOR & 0.7297  & \textbf{0.7325}    & 50.70 & \textbf{54.93} &  \textbf{0.7602} & 0.7557
 \\ 
 CRA & 0.7056  & \textbf{0.7058}    & \textbf{49.83} & 47.83 & \textbf{0.7424} & 0.7105 
 \\ 
 DEP & \textbf{0.7837}  & 0.7721    & 58.86 & \textbf{60.76} & 0.8069 & \textbf{0.8279}
 \\ 
 GWW & -       & -         			& 36.24 & \textbf{37.70} & 0.6762 & \textbf{0.6806}
 \\ \hline
 Aver. & \textbf{0.7312} & 0.7272   & 49.36 & \textbf{49.81} & \textbf{0.7368} & 0.7343
 \\ \hline 
 \end{tabular} }}
\vspace{0.2cm}
\caption{Evaluation results of the proposed multi-task SUSiNet for the three different tasks over the \mbox{COGNIMUSE} database. We report results for each movie independently as well as the average performance for each task.}
\label{table:cogn_eval}
 \end{table}

\begin{figure}[t]
\begin{center}
\hspace{-1cm}
\includegraphics[width=0.35\textwidth]{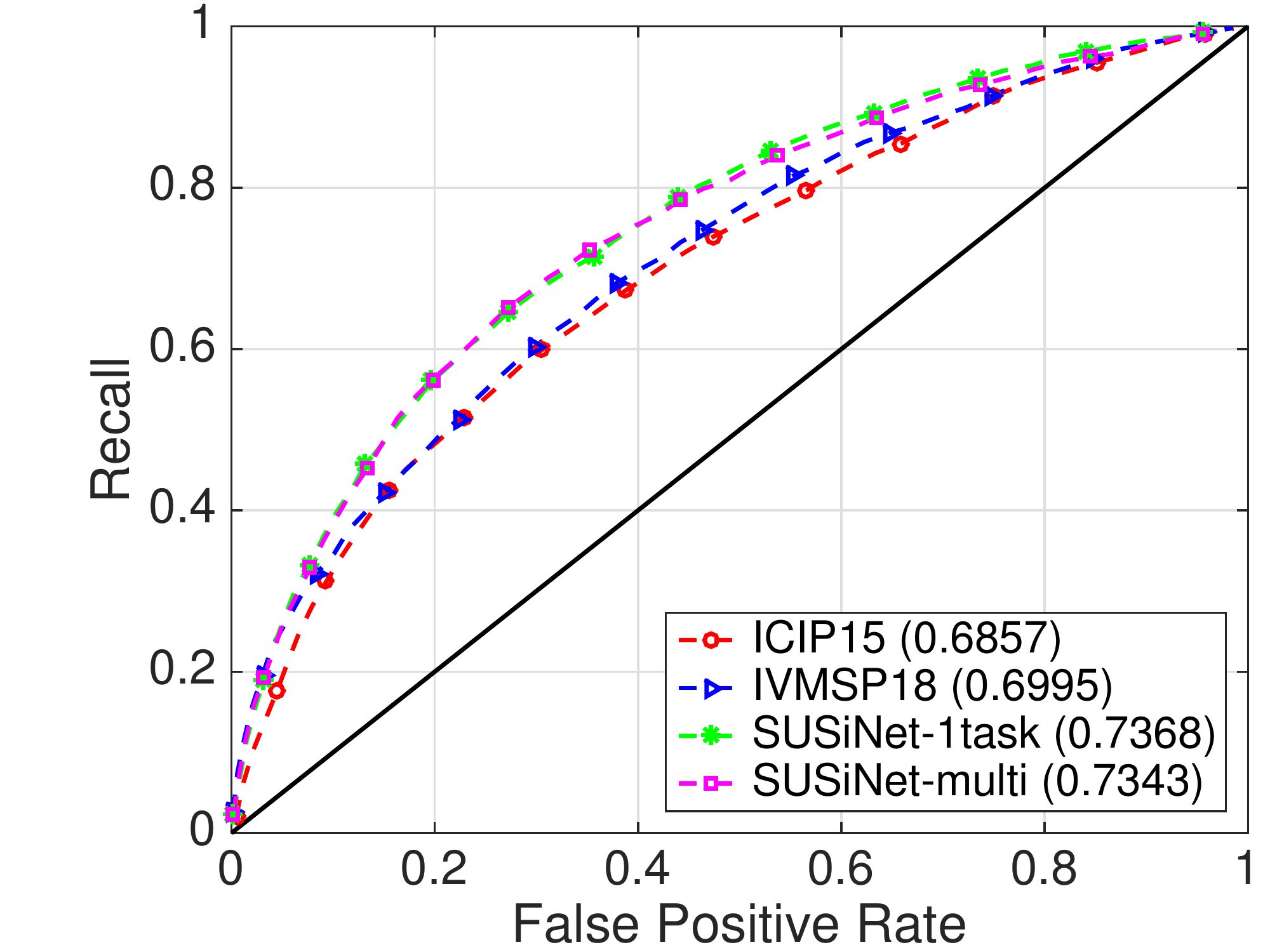}
\end{center}
\vspace{-0.4cm}
   \caption{ROC curves and corresponding AUC scores for the different methods on \mbox{COGNIMUSE} database (summarization task).}
\label{fig:roc_cogn}
\vspace{-0.3cm}
\end{figure}

\noindent\textbf{Video Summarization Evaluation:} For the evaluation over the SumMe and TVSum50 datasets we have employed the evaluation protocol of \cite{zhang2016video} that is based on the F-score between a generated keyshot-based summary (shots are temporally segmented using KTS \cite{potapov2014category}), with length 15\% of the original video duration, and the user created summaries. In Table~\ref{table:sum_eval}, we present the evaluation results for the summarization task over the SumMe and TVSum50 datasets compared against various other state-of-the-art approaches. Brackets ``[$\cdot$]" denote results obtained using an augmented dataset that includes videos from auxiliary datasets (YouTube \cite{de2011vsumm}, OVP \cite{de2011vsumm,ovp}) and it is not directly compared to our method. As we see, the multi-task SUSiNet performs very close to its single-task variant and outperforms many methods that are based on the sequential estimation of the clip based importance score, i.e., using LSTM networks. On the other hand, our network cannot perform better than other methods of  literature that operate on the whole video (i.e., using retrospective encoders \cite{zhang2018retrospective}) or employ a larger number of video frames (i.e., 128 frames in SUM-FCN \cite{rochan2018video}), especially when they are trained on the augmented dataset. However, these approaches could be added as post-hoc task-specific components and increase the performance of our network.

Regarding the evaluation of the summarization task in the \mbox{COGNIMUSE} database we have followed the evaluation procedure described in \cite{ZKE+_review}. Specifically, the summarization task is approached as a two-class classification problem, where multiple thresholds are applied to the estimated frame-wise importance scores in order to obtain results for various compression rates and thus produce summaries of various lengths. Then, the AUC metric is computed from the ROC curve that results from the different thresholds. Table~\ref{table:cogn_eval} presents the AUC scores for each movie independently as well as the average across all movies. We see that the evaluation scores of the multi- and single-task network are very close to each other, while in some cases, such us the full-length movie GWW, the multi-task network achieves better performance. It is worth to note that the low performance (around chance) for the FNE movie can be attributed to the absence of any other animation movie in the training set. As in the other cases, we have also compared our method with two other approaches that have been appeared in literature for the \mbox{COGNIMUSE} database: ICIP15 \cite{CVSP_ICIP2015} which employs hand-crafted features and IVMSP18 \cite{KZM18} that is based on C3D network. As we see in Fig.~\ref{fig:roc_cogn} the SUSiNet outperforms the two other baseline methods according to the ROC-AUC metric.

\noindent\textbf{Discussion:} In Table~\ref{table:cogn_eval} we present evaluation results for the proposed  SUSiNet in all the three different tasks over the \mbox{COGNIMUSE} database. We see that the multi-task SUSiNet, which has been jointly trained end-to-end over all tasks using very diverse datasets, can efficiently face three different problems, achieving almost the same (or in some cases better) performance to a similar single-task network, which has been trained explicitly for the specific task. In addition, as we have observed, the multi-task network performs very close to other state-of-the-art single-task methods that are based on more complex architectures or employ multiple streams networks. Figure~\ref{fig:susi_example} depicts a  qualitative example of SUSiNet on a movie excerpt from \mbox{COGNIMUSE} database, which contains sequential actions. Our network focuses on the salient regions of the video, performs action recognition and computes frame-wise importance scores simultaneously. We also observe in Fig.~\ref{fig:susi_example} that segments with the correctly recognized actions have been selected by multi-task network to be included in the video summary.     

\section{Conclusions}

In this work, we have proposed a multi-task spatio-temporal network that can jointly tackle the problems of saliency estimation, action recognition and video summarization. Our method employs only a single network for all tasks, which is jointly trained for all tasks using diverse datasets. The extensive evaluation indicates that the multi-task network performs equally well or in some cases even better than the single-task methods while it requires less computational budget than having one different network for each task. As future work, we intend to extend our network by employing more complex or multiple-streams architectures and improve the task-specific layers by incorporating recent advances from state-of-the-art single-task methods.


{\small
\bibliographystyle{ieee_fullname}
\bibliography{strings,egbib}
}

\end{document}